\documentclass[letterpaper, 10 pt, conference]{ieeeconf}  %

\IEEEoverridecommandlockouts                              %

\usepackage{graphics} %
\usepackage{times} %
\usepackage{amsmath} %
\usepackage{amssymb}  %
\usepackage{amsfonts}
\usepackage{subfigure}
\usepackage{hyperref}
\usepackage{graphicx}
\usepackage{float}
\usepackage{caption}
\usepackage{multirow}
\usepackage{booktabs}
\usepackage{color}

\title{\LARGE \bf
Differentiable Cloth Parameter Identification and State Estimation in Manipulation
}

\author{Dongzhe Zheng$^{*1}$, Siqiong Yao$^{*2}$, Wenqiang Xu$^{2}$, Cewu Lu$^{2}$%
\thanks{*Equal contribution.}%
\thanks{$^{1}$ {\tt\small dz1011@wildcats.unh.edu}. Work is done when Dongzhe is an intern at SJTU.
        }%
\thanks{$^{2}${\tt\small \{yaosiqiong, vinjohn, lucewu\}@sjtu.edu.cn}. Cewu Lu is the corresponding author, a member of Qing Yuan Research Institute and MoE Key Lab of Artificial Intelligence, AI Institute, Shanghai Jiao Tong University, Shanghai, China.
        }%
}

\begin{document}

\maketitle
\thispagestyle{empty}
\pagestyle{empty}

\begin{abstract}
In the realm of robotic cloth manipulation, accurately estimating the cloth state during or post-execution is imperative. However, the inherent complexities in a cloth's dynamic behavior and its near-infinite degrees of freedom (DoF) pose significant challenges. Traditional methods have been restricted to using keypoints or boundaries as cues for cloth state, which do not holistically capture the cloth's structure, especially during intricate tasks like folding. Additionally, the critical influence of cloth physics has often been overlooked in past research. Addressing these concerns, we introduce DiffCP, a novel differentiable pipeline that leverages the Anisotropic Elasto-Plastic (A-EP) constitutive model, tailored for differentiable computation and robotic tasks. DiffCP adopts a ``real-to-sim-to-real'' methodology. By observing real-world cloth states through an RGB-D camera and projecting this data into a differentiable simulator, the system identifies physics parameters by minimizing the geometric variance between observed and target states. Extensive experiments demonstrate DiffCP's ability and stability to determine physics parameters under varying manipulations, grasping points, and speeds. Additionally, its applications extend to cloth material identification, manipulation trajectory generation, and more notably, enhancing cloth pose estimation accuracy. More experiments and videos can be found in the supplementary materials and on the website: \url{https://sites.google.com/view/diffcp}.
\end{abstract}

\section{INTRODUCTION}
Cloth manipulation is common for humans in daily life and poses a long-standing challenge in the robotics community. When a robot performs a cloth manipulation task, it is crucial to identify whether the task is successfully executed. It requires estimating the full configuration of the cloth during or after the execution. However, estimating the cloth state from the real world is inherently challenging due to the cloth's complex dynamic behaviors and near-infinite degrees of freedom (DoF). Without the ability to estimate the full configuration, previous manipulation tasks have to be limited to keypoints (e.g., corners) or boundaries \cite{canberk2023cloth, avigal2022speedfolding} as the cues for cloth state. These cues do not represent the complete structure of the cloth, and thus cannot support to assessment of the success of more complex manipulation tasks, such as folding \cite{xue2023unifolding, avigal2022speedfolding}.

Visually tracking a highly non-rigid deformable object during manipulation is challenging. Measuring the pose directly by active sensors is not practical as discussed in ClothPose \cite{xu2023clothpose}, all known measuring techniques failed either due to accuracy or budget. Fortunately, previous works \cite{xu2023clothpose,huang2023self} have pointed out the indirect solution where the cloth is manipulated from a rather flattened state, and the partial observation of the cloth and the manipulated hand/gripper pose are recorded during the process, can measure the cloth pose. These indirect methods are purely geometry-based. However, as we know, manipulation is a dynamic process, and the deformable object is deformed due to external forces. Thus, in real-world scenarios, the influence from the physics side should also be taken into consideration for accurate cloth state estimation.

\begin{figure}[t!]
    \centering
    \includegraphics[width=1\linewidth]{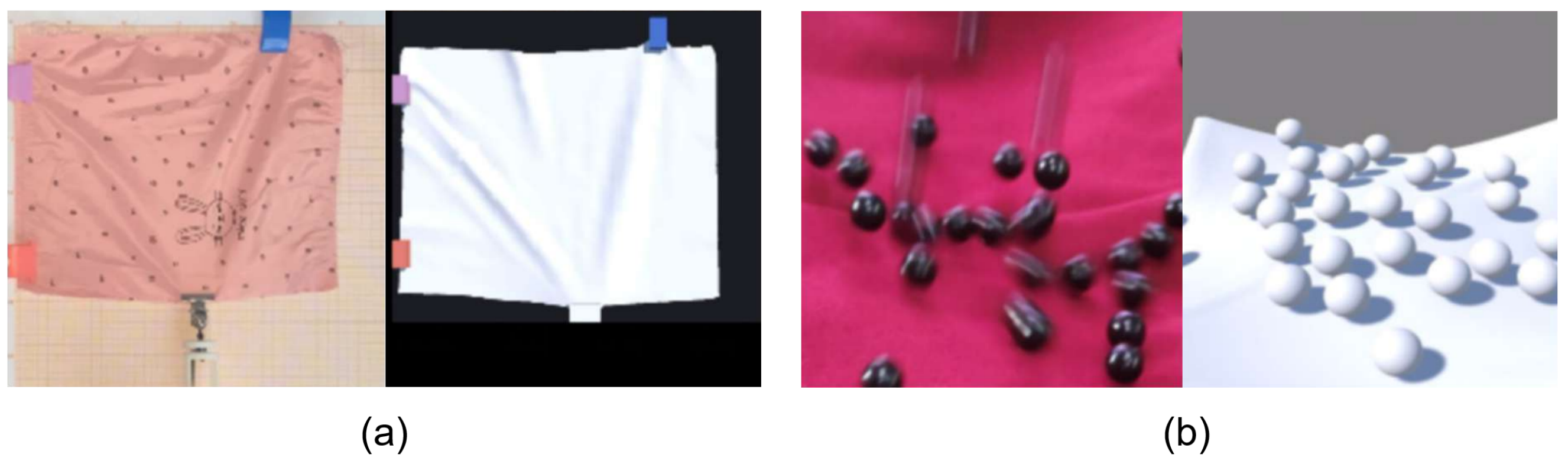}
    \caption{(a) Left: Real cloth stretching with wrinkles and folds. Right: Our A-EP simulation captures anisotropic surface details. (b) Left: Rigid spheres interacting with cloth. Right: A-EP simulation exhibits enhanced plasticity, effectively buffering and handling collisions.}
    \label{fig:property}
\end{figure}

Nevertheless, previous works on cloth manipulation \cite{clegg2018learning, chen2022efficiently} or state estimation \cite{bertiche2022neural, santesteban2019learning, lahner2018deepwrinkles} have largely overlooked the intricacies of cloth physics. Simulation-based cloth manipulation tasks \cite{canberk2023cloth} frequently resort to isotropic elastic models such as the Neo-Hookean model, which inadequately capture the true cloth behavior. As shown in Fig. \ref{fig:property}, cloths often display \textit{anisotropic elastic} properties. For instance, consider silk fabrics, which have a distinctive sheen and are more elastic along the warp threads than the weft threads. Additionally, \textit{plasticity} must also be accounted for, especially in materials with unique buffering capabilities like denim, or those with significant thickness such as woolen fabrics. Given these considerations, we adopt an Anisotropic Elasto-Plastic (A-EP) constitutive model, an approach initially introduced in the computer graphics community \cite{jiang2017anisotropic} and modify it to integrate into a differentiable pipeline and add a robot interface for robotic tasks.

Given the A-EP constitutive model, we present \textbf{DiffCP}, a differentiable pipeline to identify the cloth parameters during manipulation and make the parameters contribute to cloth pose estimation. Based on the assumption that the external force and physics parameters drive the cloth deformation, then if we match the external force between the simulation and the real world, we can estimate the physics parameters by comparing the evolving cloth states in both worlds. Specifically, DiffCP observes the cloth state in the real world with an RGB-D camera, projects the point cloud into a differentiable MPM-based simulator \cite{hu2019difftaichi}, and identifies the physics parameter by minimizing the geometric distance between the observed state and the target state. The target state can be given in either simulators or the real world.

To evaluate the proposed DiffCP, we first conduct the physics parameter identification tasks on simple-shaped fabrics and apply the estimated parameter for fabric type differentiation. Then we conduct the parameter estimation tasks on three kinds of garments with different materials and more complex shapes. We validate the stability of the estimated physics parameters with garment experiments under different manipulation operations (i.e., Cover, Fling, and Drag), different grasping points, and different operation speeds. Since the parameter estimation is carried out with a geometry-based optimization, the garment pose can be obtained along with the parameters. Furthermore, we also demonstrate the applications in manipulation trajectory generation based on the garment states.

We summarize our contribution as follows:
\begin{itemize}
    \item We propose a novel differentiable physics parameter identification framework for cloth, DiffCP. It connects the loop from real-world observation to physics simulation. We conduct extensive experiments to validate the stability of the estimated physics parameters.
    \item We apply the proposed DiffCP to three applications including cloth pose estimation, and manipulation trajectory generation. All these tasks are validated with real robot experiments.
\end{itemize}

\section{Related Works}
Our work focuses on techniques of physics parameter estimation and the applications that can be supported by the proposed frameworks such as cloth pose estimation and manipulation trajectory generation. We investigate the related works on these aspects.

\subsection{Physics Parameter Identification for Cloth}
Physics parameter estimation is an important direction in the computer graphics community. Early work \cite{param2003} estimates the dynamic simulation parameters using video sequences. However, the parameters characterize stiffness only, which is considered simplified in terms of physics.  Larionov et al. propose a differentiable framework based on position-based dynamics (PBD) \cite{param2022}. However, due to the simulation limitation of the PBD algorithm, it can only consider the elasticity of the cloth. Anatoliotakis et al. \cite{param_manip} propose to estimate the parameter of the cloth based on a mass-spring system. And the parameters to be estimated are related to the springs which means they do not have the actual physics meaning.
Our work distinguishes itself by leveraging an Anisotropic Elastoplastic (A-EP) model \cite{jiang2017anisotropic} which considers both the elasticity and plasticity of the cloth. The constitutive model is integrated within an MPM-based differentiable simulation framework, which is known to be physically accurate \cite{jiang2016material}. 

\subsection{Garment Pose Estimation}
Garment pose estimation remains a challenging task due to the high degrees of freedom associated with garment manipulation. Previous works have explored the instance-level \cite{pons2017clothcap, lahner2018deepwrinkles} and category-level \cite{chi2021garmentnets, garmenttracking, xu2023clothpose} garment pose estimation with visual input. These works can be integrated into our pipeline where we also adopt visual input of the observed point cloud. Besides, there are also works adopting tactile sensors \cite{jimenez2020perception, clegg2017learning} to estimate the state of garments. Instead of directly targeting the pose estimation, our pipeline can produce the garment pose by matching the simulated mesh to the observed point clouds to estimate the physics parameters. Thus, in this work, we just adopt the garment pose estimation results to demonstrate the stability of the physics parameter estimation, and will not benchmark the performance on this task.

\subsection{Cloth Manipulation Trajectory Generation}
Cloth manipulation in robotic applications is a long-standing research topic. Studies have explored a range of techniques, from the traditional planning \cite{hamajima1998planning, sun2014heuristic, doumanoglou2016folding} to more advanced deep reinforcement learning algorithms \cite{clegg2018learning, xue2023unifolding}, in order to facilitate the process. However, these methods use simplified cues such as keypoints and boundaries \cite{canberk2023cloth, avigal2022speedfolding} to indicate the garment states.

Recently, differentiable simulators for deformable objects \cite{gradsim,diffcloth} have been introduced in the community. In gradSim \cite{gradsim}, the cloth is modeled by the Neo-Hookean constitutive model which considers only elasticity. Besides, the cloth is controlled to a target state. However, the experiments do not converge due to the simplified state representation, center of mass. DiffCloth \cite{diffcloth} proposes a differentiable simulator based on projective dynamics. It explicitly handles the contact on the cloth, which means the manipulation trajectory will not be directly optimized but requires additional controllers. They train a neural network-based controller for hat manipulation. In contrast, since the MPM can handle the contact implicitly, besides, the manipulator and the cloth can be simultaneously taken into consideration for optimization, thus, our DiffCP can generate the manipulation trajectory directly.

\section{The Anisotropic Elastoplastic Constitutive Model for Cloth}

Inspired by the framework presented by Jiang et al. \cite{jiang2017anisotropic}, we treat cloth as a thin-shell conformal manifold. This model is characterized by two distinct energy density functions to capture strains in both the manifold and its perpendicular directions.

For strains within the manifold, our model employs a fixed corotated formulation. The associated energy function is given by:

\begin{equation}
W(F_E) = \mu \sum_{i=1}^{2}(\sigma_i-1)^2 + \frac{\lambda}{2} (J-1)^2
\end{equation}

Here, \( F_E \) signifies the deformation gradient in the manifold direction. The singular values of \( F_E \) are represented by \( \sigma_i \), while \( J \) is the determinant of \( F_E \). The parameters \( \mu \) and \( \lambda \) are the Lame constants, which dictate the elastic response of the material. These constants can be derived from the material's elastic modulus, \( E \) (a.k.a Young's modulus), and the Poisson's ratio, \( \nu \), as:

\begin{equation}
{\mu} = \frac{E}{2(1 +  \nu)}
\end{equation}
\begin{equation}
\lambda = \frac{E \nu}{(1 +  \nu)(1 - 2 \nu)}
\end{equation}

Consider an elastoplastic constitutive model orthogonal to the surface, the elastic potential energy is invariant under any rotation in the physical space, allowing us to choose a computational basis without affecting the potential energy's value. An orthogonal basis, \( Q \), is derived from the post-elastic deformation material direction, \( F_{E}D_i \), using the Gram-Schmidt process. This can also be expressed as a QR decomposition: \( QR = F_{E}D \), where \( Q \) is orthogonal and \( R \) is upper triangular. Using this basis, the elastic potential energy is defined as:

\begin{equation}
W(F_E, D) = \hat{W}(R)
\end{equation}

where 
\( D \) is the local material direction.
We split \( R \) into \( R_1 \), \( R_2 \), and \( R_3 \), representing surface deformation, surface normal shear, and surface normal compression/elongation, respectively. The energy density function is the sum of these three parts:

\begin{equation}
\hat{W}(R) = f(R_3) + g(R_2) + h(R_1)
\label{eq:energy}
\end{equation}

The component functions are defined as:

\begin{equation}
f(R_3) = \begin{cases}
k/3 (1 - r_{33})^3 & \text{if } r_{33} < 1 \\
0 & \text{if } r_{33} \geq 1
\end{cases}
\end{equation}
\begin{equation}
g(R_2) = \gamma(r_{13}^2 + r_{23}^2)
\end{equation}
\begin{equation}
h(R_1) = W(F_E)
\end{equation}

The elements \( r_{33} \), \( r_{23} \), and \( r_{13} \) of \( R_3 \) correspond to the material's elongation/compression in the normal direction and shears along the second and first directions due to deformation in the third direction, respectively.

To characterize cloth behavior under contact, the Drucker-Prager \cite{drucker1952soil} and Mohr-Coulomb models \cite{coulomb1773essai} guide our yield criterion:

\begin{equation}
||F_E - I|| \leq c_F
\end{equation}
\begin{equation}
Y(F_E) = ||F_E - I|| - c_F
\end{equation}

Here, \( I \) is the identity matrix, and \( c_F \) represents the friction coefficient. To restore the elastic state after plastic deformation, the deformation gradient considering contact \( F \) is obtained through the return mapping:

\begin{equation}
F = R(F_E) = \frac{c_F}{||F_E - I||}(F_E - I) + I
\end{equation}

\begin{figure}[t!]
    \vspace{0.3cm}
    \centering
    \includegraphics[width=1\linewidth]{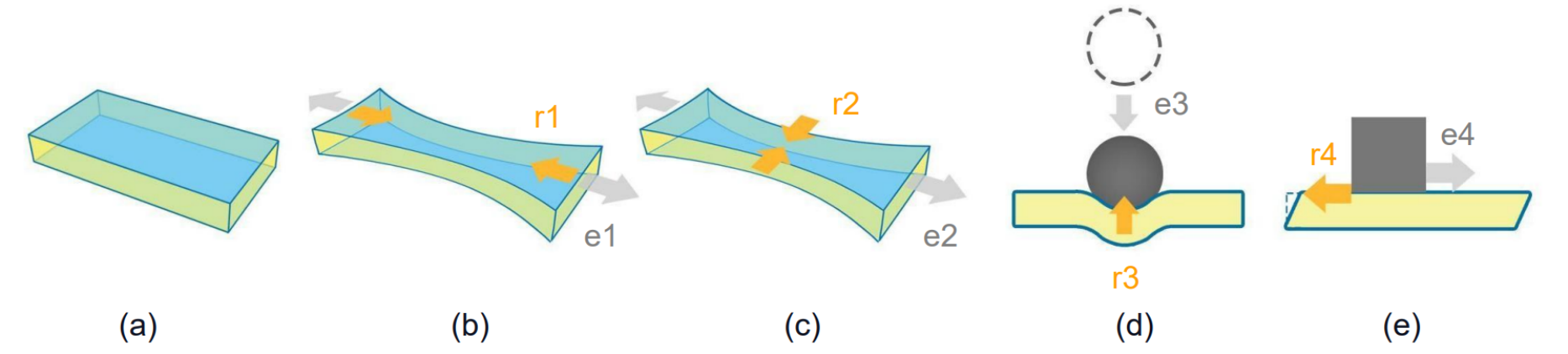}
    \caption{Five diagrams showcasing the influence of cloth physics parameters in simulations. Grey arrows represent strain (e1 to e4) on the blue cloth manifold and its pale yellow orthogonal space. Orange arrows depict the stress responses (r1 to r4) in these areas.}
    \label{fig:constitutive}
\end{figure}

Fig. \ref{fig:constitutive}a introduces cloth as a conformal manifold, capturing its thin-shell dynamics. Within, the blue intra-manifold handles distortions (Fig. \ref{fig:constitutive}b,c), while the pale yellow orthogonal space manages strains from friction and contact (Fig. \ref{fig:constitutive}d,e).

Fig. \ref{fig:constitutive}b showcases how increasing Young's modulus enhances cloth's tensile resistance along direction e1, reflected by stress r1. Fig. \ref{fig:constitutive}c ties a higher Poisson's ratio to increased resistance (r2) when stretched along e2, maintaining surface area. 

In the orthogonal space, Fig. \ref{fig:constitutive}d emphasizes the amplified cloth resistance (r3) with increased contact stiffness against force e3. Fig. \ref{fig:constitutive}e highlights the role of shear stiffness, where cloth counters shear stress (e4) with resistance (r4), preserving its geometry.

\begin{figure*}[t!]
\vspace{0.3cm}
\centering
\includegraphics[width=0.8\linewidth]{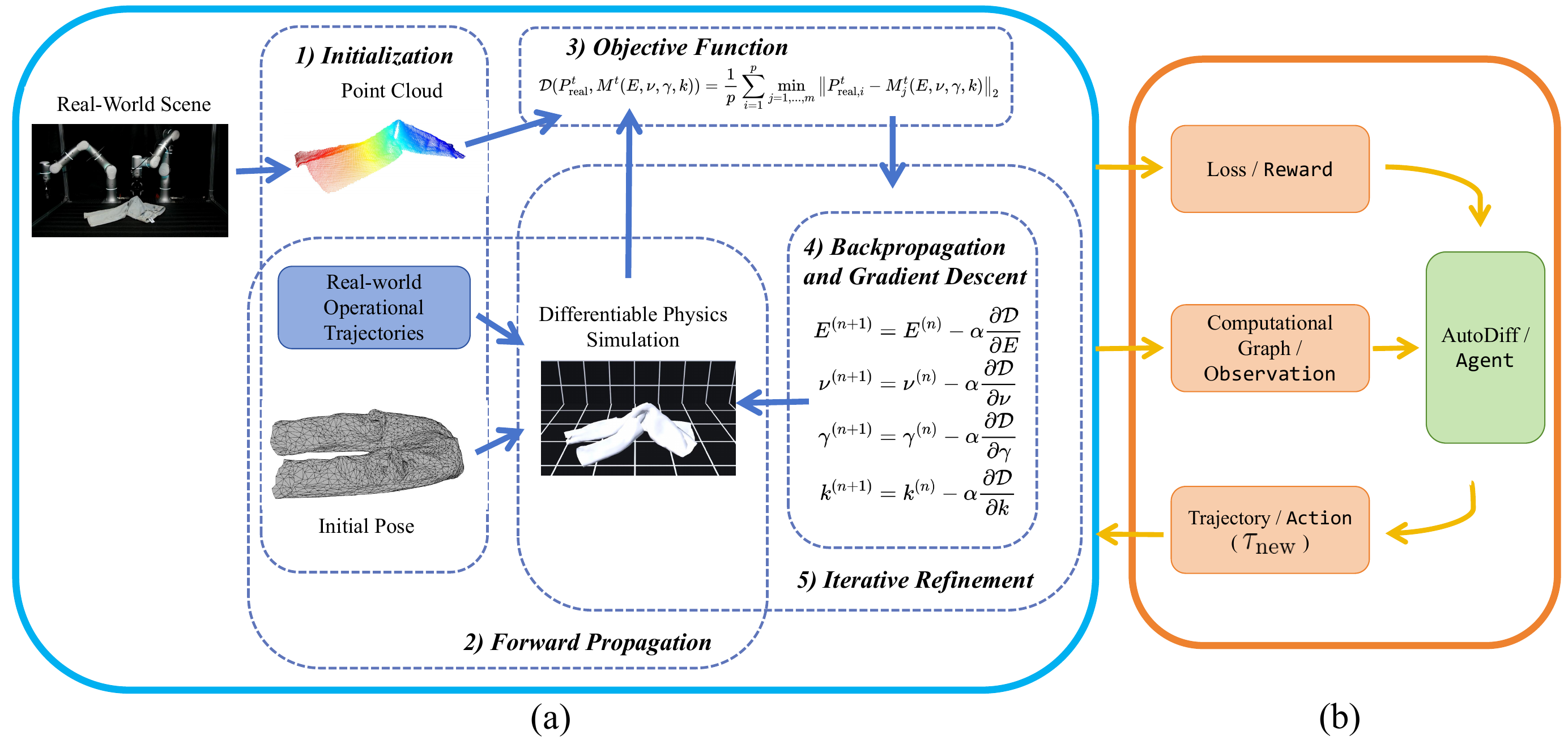}
 \caption{(a) DiffCP initializes cloth operation configurations, performs differential simulations, evaluates discrepancies via point cloud objectives, and iteratively refines cloth dynamics estimation using automatic differentiation and backpropagation. (b) With the estimated physics parameters and the differentiable simulation system, we can generate 
the manipulation trajectory by different approaches such as gradient-based optimization (GD) and reinforcement learning (RL). Components related to RL: agent, observation, and reward are distinguished using the Consolas font; GD: computational graph, loss, and gradients. Visualization uses orange for information flow, green for core algorithms, and blue arrows for simulation configurations.}
\label{fig:flowchart}
\end{figure*}

\section{DiffCP Method}
Based on the A-EP constitutive model, we propose a vision-guided differentiable parameter estimation pipeline, DiffCP. It seeks to find the constitutive parameters \( A = (E, \nu, \gamma, k) \) of the cloth in a manipulation process. The overall pipeline is given in Fig. \ref{fig:flowchart}.

\subsubsection{Initialization}
We first create a 3D model for the cloth to be manipulated by CLO software\footnote{https://www.clo3d.com/}.
Then we place the cloth model in the simulated robotic setting with an initial pose \( M^{0}(E, \nu, \gamma, k) \) and replicate the placement in the real world. Such alignment is achieved by measuring the distance from the corners of the cloth to the robot arm bases. The initial pose is flattened as far as a human judge feels satisfactory. Then, we assign an initial parameter set \( A^{(0)} \) to the cloth model. 

\subsubsection{Forward Propagation}
The robot arms are programmed to manipulate the cloth, and the operational trajectories are recorded. The trajectories are passed to the simulator and the virtual arms will move accordingly. It yields the cloth mesh vertex sequence \( M^{t}(E, \nu, \gamma, k) \). The mesh vertex sequence is evolved with the differentiable material point method (MPM) which guarantees the gradient flow during simulation. The detailed dynamic equations for vertex position update can be referred to in the supplementary materials.

\subsubsection{Objective Function}
The objective function $\text{Loss}(A) $, representing the Chamfer distance $\mathcal{D}(\cdot)$, quantifies the deviation between the simulated mesh $ M^{t}(E, \nu, \gamma, k) $ and the depth camera-based cloth dynamics captured as point cloud, which is $ P^{t}_{\text{real}} $:
\begin{align}
\text{Loss}(A) &= \mathcal{D}(P^{t}_{\text{real}}, M^{t}(E, \nu, \gamma, k)) \\
&= \frac{1}{p} \sum_{i=1}^{p} \min_{j=1,...,m} \big\| P^{t}_{\text{real}, i} \nonumber \\
&\quad - M^{t}_j(E, \nu, \gamma, k) \big\|_2
\end{align}
Here, \( m \) and \( p \) represent the total vertices in the simulated mesh and points in the real-world point cloud, respectively.

\subsubsection{Backpropagation and Gradient Descent}
Gradients are computed as:

\begin{equation}
A^{(n+1)} = A^{(n)} - \alpha \frac{\partial \text{Loss}}{\partial A}
\end{equation}
where \( \alpha \) is the learning rate and \( n \) the iteration epoch.
For the detailed formulation of the loss gradient please refer to the supplementary materials.

\subsubsection{Iterative Refinement}
Iterations persist until reaching a predefined stopping criterion, leveraging the Chamfer Distance for accuracy. The refinement stops after 35 iterations or when the Chamfer Distance is below 1.5cm.

\section{Experimental Setup}
\subsection{Clothing Items} 
In the experiments, we adopt 5 pieces of fabric: Cotton (A), Artificial Silk (B), Wool (C), Polyester (D), and Coated Nylon (E) as shown in Fig. \ref{fig:fabrics}. Besides, we also adopt 3 garments with 1 shirt, 1 pants, and 1 dress. Each with distinct materials and geometric characteristics. For the specification of the clothing items please refer to the supplementary materials. We manually build the cloth model with CLO software. To note, if the number of clothing items is bigger, we recommend using the scanning setup similar in \cite{xu2023clothpose}.

\subsection{Task Settings} 
To demonstrate the usability of the proposed DiffCP, we conduct three kinds of tasks: (1) physics parameter identification for fabrics (\textit{Fabric Experiments}). The estimated parameters can be applied to the fabric type differentiation. (2) physics parameter identification for garments (\textit{Garment Experiments}). Garments generally have more complex structures than fabrics. Thus pose estimation becomes an interesting task for garments. (3) manipulation trajectory generation. Finally, we conduct the ablation study to validate the stability of our method.

\subsubsection{Fabric Experiments}
\begin{figure}[t!]
\vspace{0.2cm}
    \centering
    \includegraphics[width=0.9\linewidth]{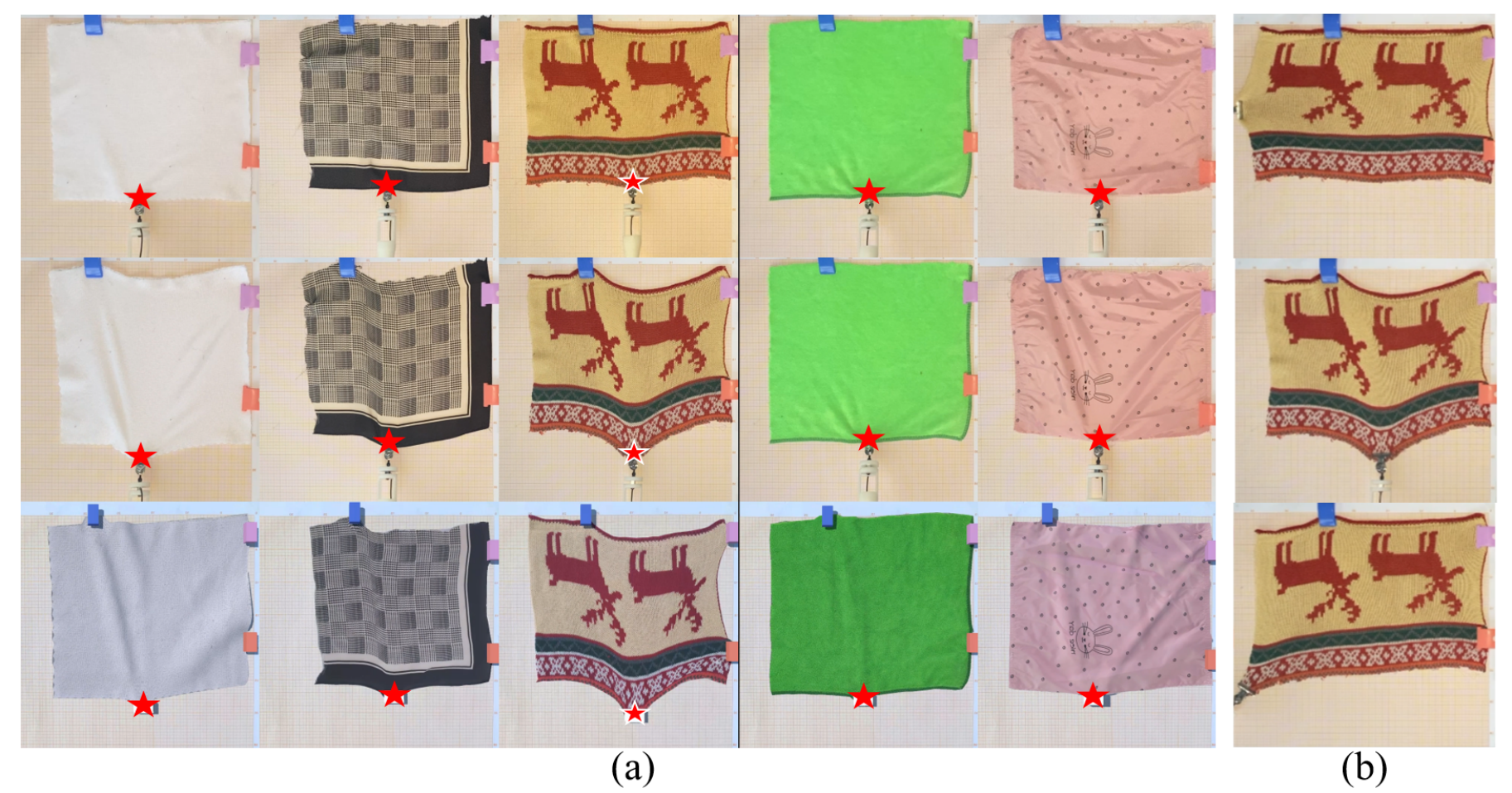}
    \caption{Fabric Manipulations for A-E: Top row depicts the initial fabric shape in real-world scenarios, the middle row shows the real-world shape post-stretching, and the bottom row presents the simulated shape after the same stretching with identified parameters. Red stars mark key points on each fabric for comparing real-world and simulation displacements: Fabric A (1.5 real, 1.75 sim cm), Fabric B (2.5 real, 2 sim cm), Fabric C (5.5 real, 5.25 sim cm), Fabric D (1 real, 1.25 sim cm), and Fabric E (0.5 real, 0.75 sim cm).
    (b)  Real-World Stretching Variations: Panels display distinct stretching directions as elaborated in the text.}
    \label{fig:fabrics}
\end{figure}

We conducted fabric experiments on rectangular cut samples.
\paragraph{Setting} We stretch the fabrics with measured forces to derive parameters: Young's modulus ($E$), Poisson's ratio ($\nu$), contact stiffness ($k$), and shear stiffness ($\gamma$). We fix the fabric to a flatten surface where a coordinated paper is placed onto the surface. We stretch the fabric with a clamp and attach a force sensor Arizon 0309 to measure the force. We vary the stretching point selection and stretching direction in the experiments. As shown in Fig. \ref{fig:fabrics}b): along x and y-axes, and at a 45-degree oblique.

\paragraph{Metrics}
Assessment metrics, including standard deviation and coefficient of variation (CoV), quantify the variations in physics parameters such as Young's modulus across different fabrics.

\subsubsection{Garment Experiments}For garments, we first identify the physical parameters during diverse actions, and then use the estimated parameters to recover the garment pose.

\begin{figure}[htbp]
    \vspace{0.2cm}
    \centering
    \includegraphics[width=0.9\linewidth]{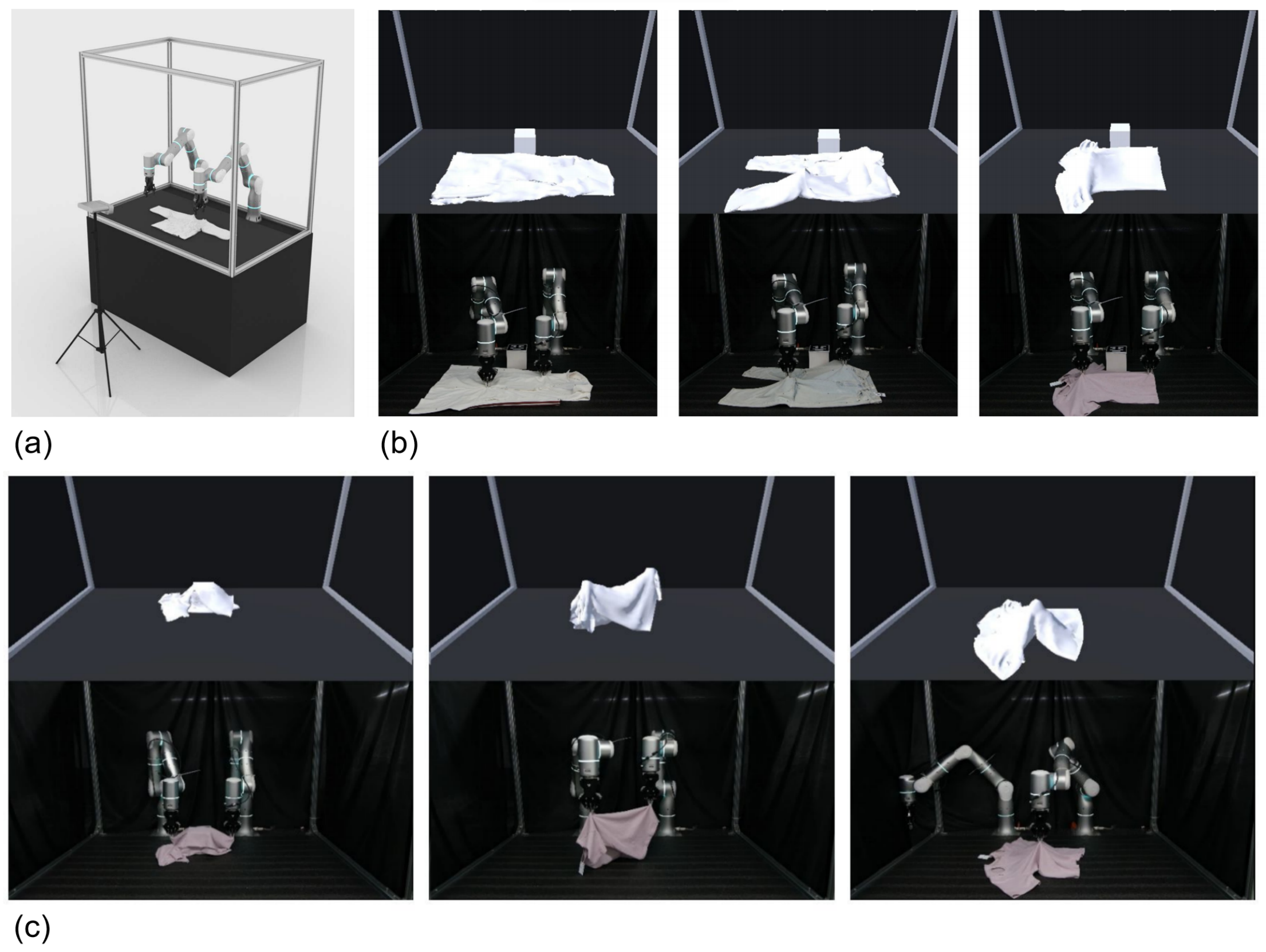}
    \caption{(a) Experimental setup showcasing two Flexiv Rizon robotic arms, integrated with an Azure Kinect depth camera for precise garment tracking. (b) Comparative panels illustrating the alignment of simulated scenarios (top) with their corresponding real-world counterparts (bottom) for three garments: dress, pants, and shirt. (c) Comparison of post-parameter identification simulated (top) with real-world (bottom) garment poses during tasks.}
    \label{fig:setup}
\end{figure}

\paragraph{Setting} The robot is programmed to perform three tasks: \textit{Cover}, \textit{Fling}, and \textit{Drag}, as illustrated in Fig. \ref{fig:setup}c. We adopt a bimanual robot setup with Flexiv Rizon robot arm the same as in \cite{xue2023unifolding}.
Deploying a robotic arm (see Fig. \ref{fig:setup}a), garment dynamics is captured using an Azure Kinect depth camera placed in the front, yielding point cloud data and concurrently logging force and grip trajectories.

Within our simulation environment (Fig. \ref{fig:setup}b), we set up a consistent representation relative to real-world scenes. Garments are modeled using 1500 triangles. For computation stability, the simulation space is downsized by 0.5x the size of the real-world counterpart, adjusting masses but maintaining densities.

\paragraph{Metrics} For physics parameters, we use CoV to evaluate the stability. For garment pose, we calculate the Chamfer distance between the simulated garment pose and the real-world observations. To gauge the stability of clothes pose estimation across conditions such as ``Action Speed'', ``Grasp Point'', and ``Initial Posture'', we compute the grand mean and pooled variance for Chamfer distance data:

\[
V_{\text{pooled}} = \frac{1}{N_{\text{total}}} \sum_{i=1}^{n} \left( N_i \cdot V_i + N_i \cdot (a_i - a_{\text{avg}})^2 \right)
\]

Where \(N_{\text{total}}\) represents the aggregate data points across all samples, \(n\) is the number of samples, \(N_i\) denotes the count of the \(i\)-th sample, \(V_i\) is its variance, \(a_i\) its average value, and \(a_{\text{avg}}\) signifies the aggregate average of all data points across samples.

\subsubsection{Manipulation Trajectory Generation} Given the differentiable property of the DiffCP system, and the estimated physics parameter, we can generate the manipulation trajectory with the input of the current state and target state. %

\begin{figure}[htbp]
\vspace{0.25cm}
    \centering
    \includegraphics[width=1\linewidth]{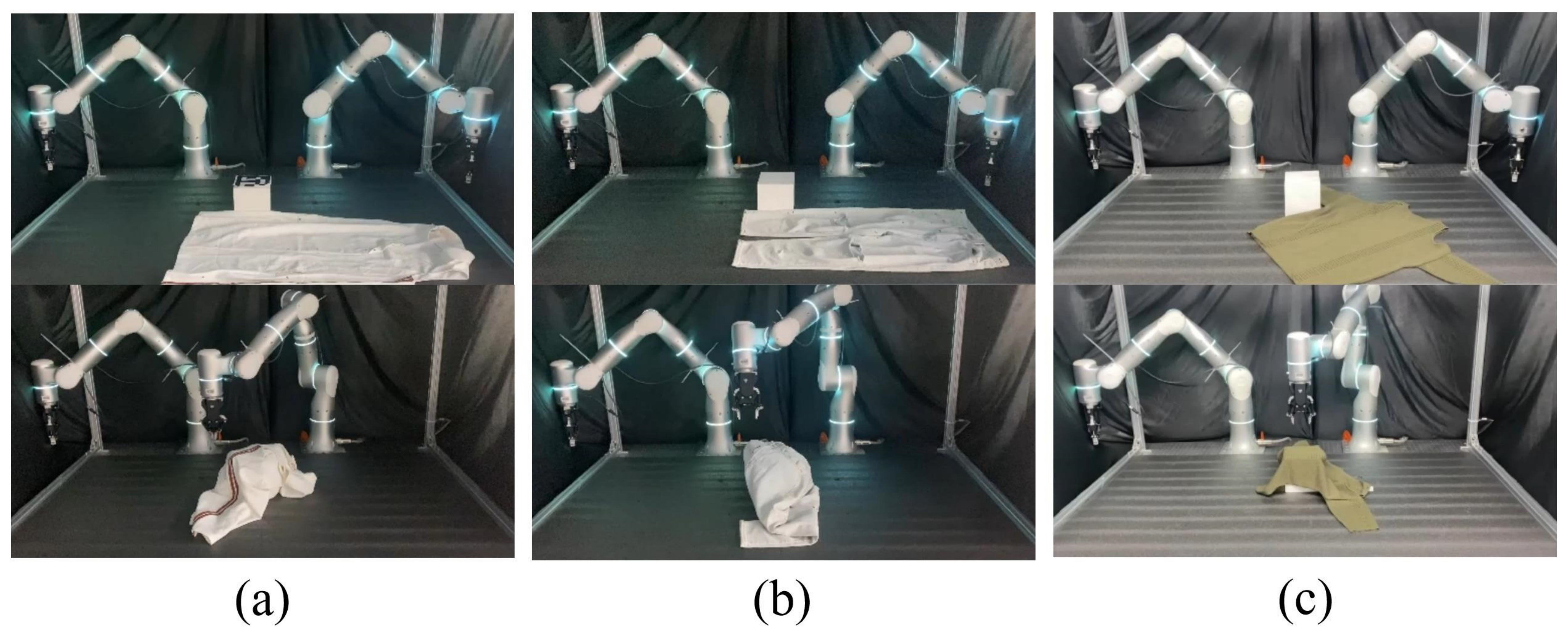}
    \caption{Experimental task setup for trajectory generation illustrated by initial and target garment postures. Three cases are presented: (a) Dress, (b) Pants, and (c) Shirt. For each case, the upper portion of the subfigure represents the initial posture, while the lower portion depicts the target posture.}
    \label{fig:trajectory_cases}
\end{figure}

\paragraph{Setting} We use Azure Kinect camera to capture the garment point cloud \(T \in \mathbb{R}^{l \times 3}\) for the current state, the target state is specified in the real world as shown in Fig. \ref{fig:trajectory_cases}. The garment is represented with triangular meshes, while the robotic arm and the rigid body are modeled as particles. The trajectory of the robotic arm is defined as \(\tau = (x_1, x_2, ..., x_L) \in \mathbb{R}^{L \times 3}\ \).
A loss function \(\mathcal{L}: \mathbb{R}^{L \times 3} \to \mathbb{R}\) gauged post-trajectory point cloud discrepancies, optimizing through the gradient function \(\nabla_\tau \mathcal{L}\). The loss function employs the Chamfer distance to directly measure the disparity between the manipulated cloth's shape and the target. Minimizing this distance inherently guides the trajectory of the grasp points, ensuring a path that results in a shape closely resembling the desired target.

\paragraph{Baseline method for trajectory generation}
We adopt a reinforcement learning method as the trajectory generation baseline. We use the Soft Actor-Critic algorithm \cite{sac}, with the state space \(\mathcal{S}\) and action space \(\mathcal{A}\). The reward is defined by the negative Chamfer distance between the observed garment state and the target state. The training details can be referred to in the supplementary materials.

\paragraph{Metrics}
We calculate the Chamfer distance between the final garment pose and the target pose.

\subsubsection{Parameter Ablation Analysis}

In this ablation analysis, we scrutinize the role and impact of specific parameters within the dynamics of the garment. %

\paragraph{Setting}
We use the three garments in the ablation study. All the garments are manipulated by three actions: Cover, Fling, and Drag. Since our aim is to validate the importance of the parameters, the experiments can all be conducted in the simulators after the reference values of the parameters are estimated by the real-sim-real loop of the identification process.

Each estimated parameters are scaled at six levels: 25\%, 50\%, 75\%, 125\%, 150\%, and 175\%. For the Poisson's ratio \(\nu\), which should be bounded in [0, 0.5], we employ the rescaling function \(f(x)\) to ensure scaled values are both monotonically increasing and confined within the physical bound:
\begin{equation}
f(x) = \frac{1}{0.5 - x} -2
\end{equation}

\paragraph{Metrics} We use the bidirectional Chamfer distance, which is the sum of the unidirectional Chamfer distances in both directions, to compare the shapes of cloth simulations with rescaled parameters versus original parameters.

Besides, to calculate the importance of the parameters, we adopt an index of energy contribution which is defined as:
\begin{equation}
C_i^j = \frac{W_i - W_i^j}{W_i}
\end{equation}
where \(i\) signifies the action, and \(j\) indicates the parameter. Derived from the energy function in Equation \ref{eq:energy}, \(W_i\) signifies the total energy, whereas \(W_i^j\) captures the energy with parameter \(j\) excluded.

\paragraph{Discussion: Baseline method for physics parameter identification} It is noteworthy we do not specify a baseline method for physics parameter identification. Because, to the best of our knowledge, we have not found a previous method that properly fits ours as a baseline.  For the constitutive model, when we ablate the contact stiffness \(k\), and shear stiffness \(\gamma\), it is equivalent to the fixed corotated formulation. We consider it as the baseline approach.

\section{Results}

\subsection{Fabric Experiments}

\subsubsection{Physics Parameter Identification} We conduct 10 times of the stretching with different grasping points and directions, estimate the parameters each time, average them and report in Table \ref{table:mean_parameters_of_fabrics}.

\begin{table}[htbp]
\vspace{0.25cm}
\centering
\caption{Mean parameters of different fabrics.}
\begin{tabular}{ccccc}
\toprule
Fabric & $E$ & $\nu$ & $k$ & $\gamma$ \\
\midrule
Cotton (A) & 609.02 & 0.15 & 6520.57 & 1674.00 \\
Artificial Silk (B) & 713.00 & 0.37 & 9330.89 & 553.84 \\
Wool (C) & 221.37 & 0.31 & 4729.10 & 2000.57 \\
Polyester (D) & 1792.45 & 0.13 & 8715.03 & 1260.79 \\
Coated Nylon (E) & 3579.50 & 0.06 & 9638.63 & 2388.09 \\
\bottomrule
\end{tabular}
\label{table:mean_parameters_of_fabrics}
\vspace{-0.25cm}
\end{table}

\subsubsection{Coefficient of Variation Analysis}
To assess the precision and repeatability of our measurements across these fabrics, we analyzed the coefficient of variation (CoV), summarized in Table \ref{table:cv_for_fabrics}.

\begin{table}[htbp]
\centering
\caption{Coefficient of Variation (CoV) for Different Fabrics}
\begin{tabular}{ccccc}
\toprule
Fabric & $E$ (\%)  & $\nu$ (\%) & $k$ (\%) & $\gamma$ (\%) \\
\midrule
Cotton (A) & 0.384 & 6.0 & 4.109 & 8.498 \\
Artificial Silk (B) & 0.369 & 1.081 & 6.148 & 6.146 \\
Wool (C) & 7.932 & 3.548 & 5.940 & 17.829 \\
Polyester (D) & 0.461 & 4.615 & 2.252 & 10.902 \\
Coated Nylon (E) & 0.859 & 10.0 & 13.823 & 3.747 \\
\bottomrule
\end{tabular}
\label{table:cv_for_fabrics}
\end{table}

Our inter-fabric variability analysis showed most parameters consistently below a 1\% CoV, highlighting our method's precision across fabrics. %

Artificial Silk's minimal variability emphasizes its inherent stability. In contrast, Wool, with its anisotropic nature due to unique weaving, presented wider measurement ranges. Coated Nylon's stiffness, attributed to its coating, led to minimal deformations in simulations. Such limited deformations amplify even minor measurement errors, resulting in its higher variability.

While $E$ and $\nu$ exhibited minimal variations, contact stiffness ($k$) and shear stiffness ($\gamma$) displayed greater fabric-specific variations, aligning with our experimental design, since the former two are considered more important.

\subsection{Garment Experiments}

\subsubsection{Physics Parameter Identification}

Similarly, we conduct 10 times manipulation on all three garments by grasping and placing them from different grasping points, then we obtain the average and the CoV of the estimated parameters and report in Table~\ref{combined_data}.

\begin{table}[htbp]
\centering
\caption{Mean Values with Coefficient of Variation (CoV) for Garment Parameters}
\begin{tabular}{lccc}
\toprule
Parameter & Dress & Pants & Shirts \\
\midrule
\( E \) & 129.32 (11\%) & 638.73 (8.4\%) & 328.77 (8.4\%) \\
\( \nu \) & 0.121 (10.7\%) & 0.071 (9.9\%) & 0.103 (10.7\%) \\
\( k \) & 5495.34 (11.4\%) & 9498.36 (7.2\%) & 2301.56 (12.3\%) \\
\( \gamma \) & 350.88 (18.6\%) & 733.46 (16.7\%) & 149.43 (24.8\%) \\
\bottomrule
\end{tabular}
\label{combined_data}
\end{table}

Despite the inherent differences in the physical properties across garments, our parameter estimation technique manifests robustness and reliability.

\subsubsection{Pose Estimation and Stability Validation}
During the simulation process, we can naturally obtain the garment pose. We report the Chamfer distance between the garment mesh in the simulator and the observed point clouds from the real world. It shows the quality of the garment pose derived from the simulation. Besides, by utilizing the Chamfer distance, we can also evaluate the factors that can influence the simulation accuracy, resulting in garment pose accuracy. The factors are action types (reported in Table \ref{chamfer_data_tasks}) and garment types (reported in Table \ref{chamfer_data_clothes}). The data is collected by 10-time experiments.

\begin{table}[htbp]
\centering
\caption{Chamfer Distance for Different Actions.}
\begin{tabular}{lcc}
\toprule
Action & Mean Chamfer Distance & \(CoV\) (\%) \\
\midrule
Cover & 0.017143 & 21.37\% \\
Drag & 0.016957 & 23.16\% \\
Fling & 0.014176 & 30.28\% \\
\bottomrule
\end{tabular}
\label{chamfer_data_tasks}
\vspace{-0.5cm}
\end{table}

\begin{table}[htbp]
\centering
\caption{Chamfer Distance for Different Garments.}
\begin{tabular}{lcc}
\toprule
Garment & Mean Chamfer Distance & \(CoV\) (\%) \\
\midrule
Dress & 0.023448 & 17.09\% \\
Pants & 0.015000 & 37.01\% \\
Shirt & 0.009829 & 23.63\% \\
\bottomrule
\end{tabular}
\label{chamfer_data_clothes}
\vspace{-0.2cm}
\end{table}

\subsubsection{Influence of System Variables} We also consider whether other factors will influence the pose estimation, such as ``Action Speed'', ``Grasp Point'', ``Initial Posture'' of garments.

The grand mean for ``Action Speed'' is measured at 0.0160 with an associated pooled variance of \(7.14 \times 10^{-5}\). For the ``Grasp Point'', the grand mean is slightly higher at 0.0162, accompanied by a pooled variance of \(8.74 \times 10^{-5}\). Lastly, the ``Initial Posture'' yieldes a grand mean of 0.0161 and has a pooled variance of \(8.12 \times 10^{-5}\). Among these variables, the ``Grasp Point'' exhibits the most variability, suggesting its influence on the simulation outcomes.

\subsection{Manipulation Trajectory Generation}
For manipulation trajectory generation, we compare the direct optimization (denoted as ``GD'') from the differentiable simulator with the reinforcement learning baseline (denoted as ``RL'').

\begin{figure}[t!]
\vspace{0.2cm}
    \centering
    \includegraphics[width=1\linewidth]{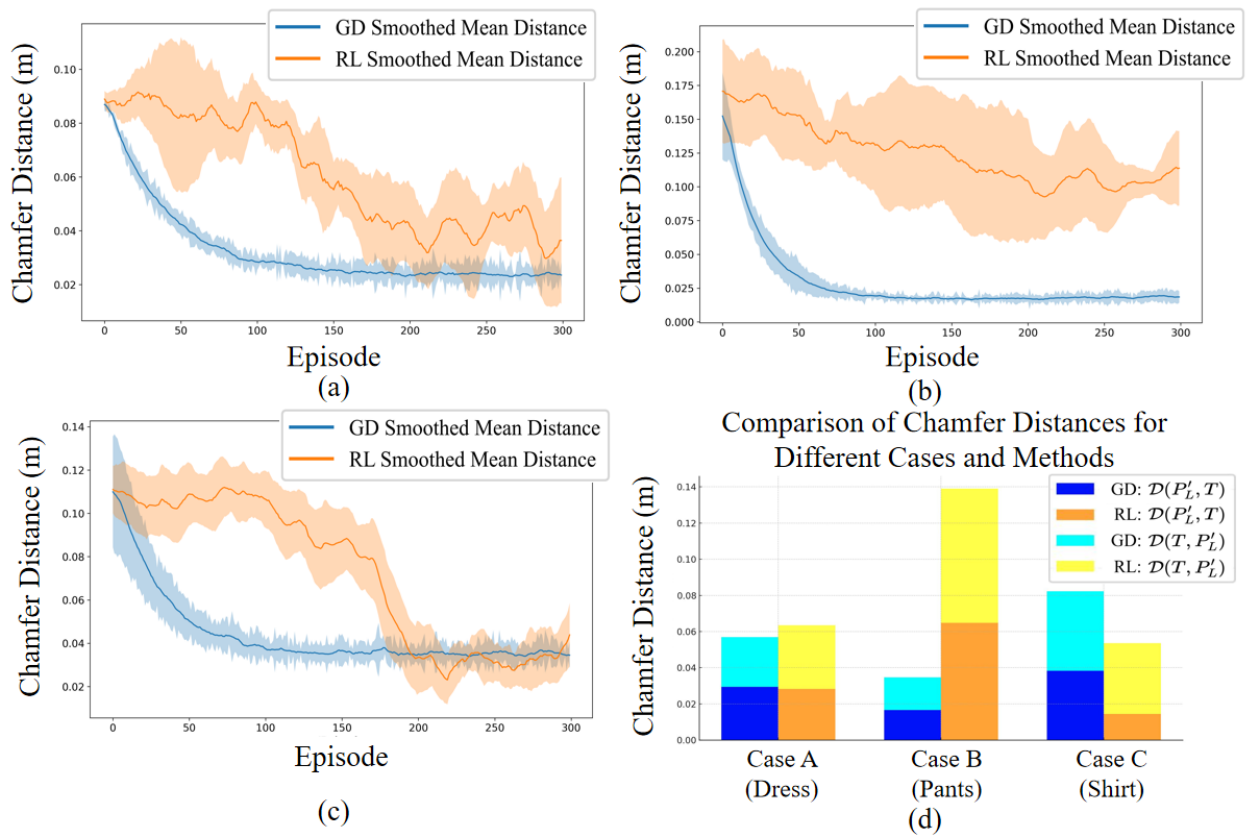}
    \caption{(a-c) Simulation Convergence for Dress (Case A), Pants (Case B), and Shirt (Case C) respectively: Chamfer distances plotted for RL and GD methods across successive episodes. (d) Real-world Distance Comparison for Dress, Pants, and Shirt.}

    \label{fig:traj_gen_results}
    \vspace{-0.5cm}
\end{figure}

In Figure \ref{fig:traj_gen_results}a-c, we observe GD's inherent stability in trajectory generation. Notably, GD typically converges to an optimal trajectory in fewer episodes compared to RL. This trend is consistent across all task scenarios, from dress to shirt.
While GD displays a stable and efficient convergence in most cases, RL exhibits advantages in specific tasks. For instance, when manipulating shirts, RL achieves a marginally superior Chamfer distance, hinting at its potential in complex manipulations.

Real-world results (Figure \ref{fig:traj_gen_results}d) echo our simulation findings, further strengthening the credibility of our simulations.

\subsection{Ablation Study}

\subsubsection{Parameter Impact on Chamfer Distance}

Each parameter's perturbation led to discernible alterations in Chamfer distance (CD) across tasks, revealing the multifaceted nature of cloth simulations. Specifically, the influence of contact stiffness (\(k\)) and shear stiffness (\(\gamma\)) across all tasks underscores their indispensability, especially in tasks marked by extensive contact and collisions, exemplified by the Cover task. Notably, while the magnitude reductions in \(k\) and \(\gamma\) exerted marginal effects, their increases led to significant shifts in the simulation outcomes. In contrast, variations in \(E\) and \(\nu\) consistently show significant impacts.

\begin{table}[htbp]
\centering
\caption{Chamfer Distance across tasks for scaled parameters relative to their baseline (100\%).}
\begin{tabular}{lccccccc}
\toprule
Task/Param. & 25 & 50 & 75 & 100 & 125 & 150 & 175 \\
\midrule
Fling \(E\) & 0.72 & 0.36 & 0.16 & - & 0.09 & 0.13 & 0.15 \\
Fling \(\nu\) & 0.34 & 0.32 & 0.07 & - & 0.03 & 0.06 & 0.08 \\
Fling \(k\) & 0.01 & 0.01 & 0.00 & - & 0.01 & 0.03 & 0.04 \\
Fling \(\gamma\) & 0.02 & 0.02 & 0.01 & - & 0.02 & 0.03 & 0.03 \\
\hline
Drag \(E\) & 0.84 & 0.34 & 0.14 & - & 0.07 & 0.12 & 0.12 \\
Drag \(\nu\) & 0.24 & 0.22 & 0.07 & - & 0.03 & 0.04 & 0.06 \\
Drag \(k\) & 0.01 & 0.01 & 0.01 & - & 0.00 & 0.00 & 0.00 \\
Drag \(\gamma\) & 0.02 & 0.02 & 0.01 & - & 0.01 & 0.02 & 0.02 \\
\hline
Cover \(E\) & 0.71 & 0.36 & 0.16 & - & 0.04 & 0.05 & 0.51 \\
Cover \(\nu\) & 0.36 & 0.35 & 0.23 & - & 0.08 & 0.09 & 0.11 \\
Cover \(k\) & 0.04 & 0.01 & 0.01 & - & 0.23 & 0.51 & 0.52 \\
Cover \(\gamma\) & 0.02 & 0.02 & 0.01 & - & 0.11 & 0.24 & 0.28 \\
\bottomrule
\end{tabular}
\label{tab:CD_across_tasks}

\end{table}

\subsubsection{Parameter Contribution to Energy Density}

Delving deeper, we sought to elucidate the intricate interplay of \(E\), \(\nu\), \(k\), and \(\gamma\) in sculpting the energy density terrain across the aforementioned tasks. The analysis, as reported in Table \ref{tab:energy_density_contribution}, unravels the multifarious influences of these parameters.

The overarching dominance of \(E\), particularly in tasks characterized by superficial elastic deformations, cannot be overstated. Conversely, the roles of \(k\) and \(\gamma\) remain relatively subdued, suggesting minimal plastic deformations orthogonal to the fabric direction. Poisson's ratio \(\nu\) emerges as a pivotal parameter, elucidating the proportional relationship between the fabric's lateral and longitudinal deformations, albeit its influence remains secondary to \(E\).

\begin{table}[htbp]
\vspace{0.2cm}
\centering
\caption{Energy Density Contribution of Different Parameters across Tasks}
\begin{tabular}{lccc}
\toprule
Parameter & Fling & Drag & Cover\\
\midrule
\(E\) & 99.8\% & 99.9\% & 74.8\%\\
\(\nu\) & 15.5 \% & 23.7\% & 26.3\%\\
\(k\) & 0.4\% & 0.2\% & 12.6\%\\
\(\gamma\) & 0.2\% & 0.7\% & 8.7\%\\
\bottomrule
\end{tabular}
\label{tab:energy_density_contribution}
\vspace{-0.4cm}
\end{table}

\begin{figure}[htp]
    \centering
    \includegraphics[width=1\linewidth]{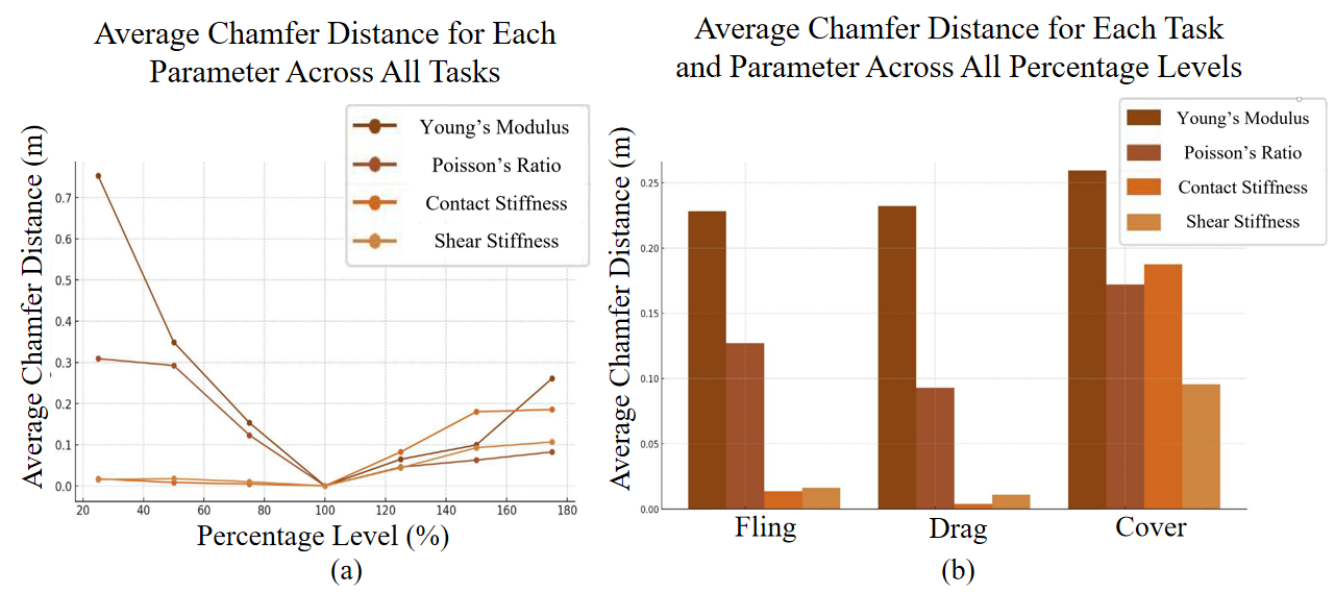}
    \caption{Analyses of Chamfer Distance variations with respect to constitutive parameter scaling. (a) Showcases how different parameters influence each task as they are scaled from 25\% to 175\%. (b) Illustrates the cumulative effect of parameter scaling on the Chamfer Distance across all tasks.}
    \label{fig:avg_cd}
    \vspace{-0.5cm}
\end{figure}

\section{Conclusion and Future Works}
In this study, we present DiffCP, a differentiable cloth simulation pipeline stemming from DiffMPM and integrated with the Anisotropic Elasto-Plastic (A-EP) model to capture intricate cloth behaviors. This enhances robot adaptability in manipulating varied cloth materials. Our experiments validate our method's precision in discerning fabric properties and estimating garment poses under robotic operations. However, challenges arise due to real-simulated data discrepancies, largely from sensor inaccuracies, and the significant memory demands of differentiable simulations. We're addressing these by refining data alignment and improving memory efficiency in future work.

\bibliographystyle{IEEEtran}
\bibliography{main}

\begin{thebibliography}{10}
\providecommand{\url}[1]{#1}
\csname url@rmstyle\endcsname
\providecommand{\newblock}{\relax}
\providecommand{\bibinfo}[2]{#2}
\providecommand\BIBentrySTDinterwordspacing{\spaceskip=0pt\relax}
\providecommand\BIBentryALTinterwordstretchfactor{4}
\providecommand\BIBentryALTinterwordspacing{\spaceskip=\fontdimen2\font plus
\BIBentryALTinterwordstretchfactor\fontdimen3\font minus
  \fontdimen4\font\relax}
\providecommand\BIBforeignlanguage[2]{{%
\expandafter\ifx\csname l@#1\endcsname\relax
\typeout{** WARNING: IEEEtran.bst: No hyphenation pattern has been}%
\typeout{** loaded for the language `#1'. Using the pattern for}%
\typeout{** the default language instead.}%
\else
\language=\csname l@#1\endcsname
\fi
#2}}

\bibitem{canberk2023cloth}
A.~Canberk, C.~Chi, H.~Ha, B.~Burchfiel, E.~Cousineau, S.~Feng, and S.~Song,
  ``Cloth funnels: Canonicalized-alignment for multi-purpose garment
  manipulation,'' in \emph{2023 IEEE International Conference on Robotics and
  Automation (ICRA)}.\hskip 1em plus 0.5em minus 0.4em\relax IEEE, 2023, pp.
  5872--5879.

\bibitem{avigal2022speedfolding}
Y.~Avigal, L.~Berscheid, T.~Asfour, T.~Kr{\"o}ger, and K.~Goldberg,
  ``Speedfolding: Learning efficient bimanual folding of garments,'' in
  \emph{2022 IEEE/RSJ International Conference on Intelligent Robots and
  Systems (IROS)}.\hskip 1em plus 0.5em minus 0.4em\relax IEEE, 2022, pp. 1--8.

\bibitem{xue2023unifolding}
H.~Xue, Y.~Li, W.~Xu, D.~Zheng, and C.~Lu, ``Unifolding: Towards
  sample-efficient, scalable, and generalizable robotic garment folding,'' in
  \emph{7th Annual Conference on Robot Learning}, 2023.

\bibitem{xu2023clothpose}
W.~Xu, W.~Du, H.~Xue, Y.~Li, R.~Ye, Y.~Wang, and C.~Lu, ``Clothpose: A
  real-world benchmark for visual analysis of garment pose via an indirect
  recording solution,'' in \emph{Proceedings of the IEEE/CVF International
  Conference on Computer Vision}, 2023.

\bibitem{huang2023self}
Z.~Huang, X.~Lin, and D.~Held, ``Self-supervised cloth reconstruction via
  action-conditioned cloth tracking,'' \emph{arXiv preprint arXiv:2302.09502},
  2023.

\bibitem{clegg2018learning}
A.~Clegg, W.~Yu, J.~Tan, C.~K. Liu, and G.~Turk, ``Learning to dress:
  Synthesizing human dressing motion via deep reinforcement learning,''
  \emph{ACM Transactions on Graphics (TOG)}, vol.~37, no.~6, pp. 1--10, 2018.

\bibitem{chen2022efficiently}
L.~Y. Chen, H.~Huang, E.~Novoseller, D.~Seita, J.~Ichnowski, M.~Laskey,
  R.~Cheng, T.~Kollar, and K.~Goldberg, ``Efficiently learning single-arm fling
  motions to smooth garments,'' in \emph{The International Symposium of
  Robotics Research}.\hskip 1em plus 0.5em minus 0.4em\relax Springer, 2022,
  pp. 36--51.

\bibitem{bertiche2022neural}
H.~Bertiche, M.~Madadi, and S.~Escalera, ``Neural cloth simulation,'' \emph{ACM
  Transactions on Graphics (TOG)}, vol.~41, no.~6, pp. 1--14, 2022.

\bibitem{santesteban2019learning}
I.~Santesteban, M.~A. Otaduy, and D.~Casas, ``Learning-based animation of
  clothing for virtual try-on,'' in \emph{Computer Graphics Forum}, vol.~38,
  no.~2.\hskip 1em plus 0.5em minus 0.4em\relax Wiley Online Library, 2019, pp.
  355--366.

\bibitem{lahner2018deepwrinkles}
Z.~Lahner, D.~Cremers, and T.~Tung, ``Deepwrinkles: Accurate and realistic
  clothing modeling,'' in \emph{Proceedings of the European conference on
  computer vision (ECCV)}, 2018, pp. 667--684.

\bibitem{jiang2017anisotropic}
C.~Jiang, T.~Gast, and J.~Teran, ``Anisotropic elastoplasticity for cloth, knit
  and hair frictional contact,'' \emph{ACM Transactions on Graphics (TOG)},
  vol.~36, no.~4, pp. 1--14, 2017.

\bibitem{hu2019difftaichi}
Y.~Hu, L.~Anderson, T.-M. Li, Q.~Sun, N.~Carr, J.~Ragan-Kelley, and F.~Durand,
  ``Difftaichi: Differentiable programming for physical simulation,''
  \emph{arXiv preprint arXiv:1910.00935}, 2019.

\bibitem{param2003}
K.~S. Bhat, C.~D. Twigg, J.~K. Hodgins, P.~Khosla, Z.~Popovic, and S.~M. Seitz,
  ``Estimating cloth simulation parameters from video,'' 2003.

\bibitem{param2022}
E.~Larionov, M.-L. Eckert, K.~Wolff, and T.~Stuyck, ``Estimating cloth
  elasticity parameters using position-based simulation of compliant
  constrained dynamics,'' \emph{arXiv preprint arXiv:2212.08790}, 2022.

\bibitem{param_manip}
N.~E. Anatoliotakis, P.~Koustoumpardis, and K.~Moustakas, ``Cloth mechanical
  parameter estimation and simulation for optimized robotic manipulation,'' in
  \emph{Proceedings of the IEEE/CVF International Conference on Computer
  Vision}, 2021, pp. 2612--2620.

\bibitem{jiang2016material}
C.~Jiang, C.~Schroeder, J.~Teran, A.~Stomakhin, and A.~Selle, ``The material
  point method for simulating continuum materials,'' in \emph{Acm siggraph 2016
  courses}, 2016, pp. 1--52.

\bibitem{pons2017clothcap}
G.~Pons-Moll, S.~Pujades, S.~Hu, and M.~J. Black, ``Clothcap: Seamless 4d
  clothing capture and retargeting,'' \emph{ACM Transactions on Graphics
  (ToG)}, vol.~36, no.~4, pp. 1--15, 2017.

\bibitem{chi2021garmentnets}
C.~Chi and S.~Song, ``Garmentnets: Category-level pose estimation for garments
  via canonical space shape completion,'' in \emph{Proceedings of the IEEE/CVF
  International Conference on Computer Vision}, 2021, pp. 3324--3333.

\bibitem{garmenttracking}
H.~Xue, W.~Xu, J.~Zhang, T.~Tang, Y.~Li, W.~Du, R.~Ye, and C.~Lu,
  ``Garmenttracking: Category-level garment pose tracking,'' in
  \emph{Proceedings of the IEEE/CVF Conference on Computer Vision and Pattern
  Recognition}, 2023, pp. 21\,233--21\,242.

\bibitem{jimenez2020perception}
P.~Jim{\'e}nez and C.~Torras, ``Perception of cloth in assistive robotic
  manipulation tasks,'' \emph{Natural Computing}, vol.~19, pp. 409--431, 2020.

\bibitem{clegg2017learning}
A.~Clegg, W.~Yu, Z.~Erickson, J.~Tan, C.~K. Liu, and G.~Turk, ``Learning to
  navigate cloth using haptics,'' in \emph{2017 IEEE/RSJ International
  Conference on Intelligent Robots and Systems (IROS)}.\hskip 1em plus 0.5em
  minus 0.4em\relax IEEE, 2017, pp. 2799--2805.

\bibitem{hamajima1998planning}
K.~Hamajima and M.~Kakikura, ``Planning strategy for task untangling
  laundry-isolating clothes from a washed mass,'' \emph{Journal of Robotics and
  Mechatronics}, vol.~10, no.~3, pp. 244--251, 1998.

\bibitem{sun2014heuristic}
L.~Sun, G.~Aragon-Camarasa, P.~Cockshott, S.~Rogers, and J.~P. Siebert, ``A
  heuristic-based approach for flattening wrinkled clothes,'' in \emph{Towards
  Autonomous Robotic Systems: 14th Annual Conference, TAROS 2013, Oxford, UK,
  August 28--30, 2013, Revised Selected Papers 14}.\hskip 1em plus 0.5em minus
  0.4em\relax Springer, 2014, pp. 148--160.

\bibitem{doumanoglou2016folding}
A.~Doumanoglou, J.~Stria, G.~Peleka, I.~Mariolis, V.~Petrik, A.~Kargakos,
  L.~Wagner, V.~Hlav{\'a}{\v{c}}, T.-K. Kim, and S.~Malassiotis, ``Folding
  clothes autonomously: A complete pipeline,'' \emph{IEEE Transactions on
  Robotics}, vol.~32, no.~6, pp. 1461--1478, 2016.

\bibitem{gradsim}
J.~K. Murthy, M.~Macklin, F.~Golemo, V.~Voleti, L.~Petrini, M.~Weiss,
  B.~Considine, J.~Parent-L{\'e}vesque, K.~Xie, K.~Erleben, \emph{et~al.},
  ``gradsim: Differentiable simulation for system identification and visuomotor
  control,'' in \emph{International conference on learning representations},
  2020.

\bibitem{diffcloth}
Y.~Li, T.~Du, K.~Wu, J.~Xu, and W.~Matusik, ``Diffcloth: Differentiable cloth
  simulation with dry frictional contact,'' \emph{ACM Transactions on Graphics
  (TOG)}, vol.~42, no.~1, pp. 1--20, 2022.

\bibitem{drucker1952soil}
D.~C. Drucker and W.~Prager, ``Soil mechanics and plastic analysis or limit
  design,'' \emph{Quarterly of applied mathematics}, vol.~10, no.~2, pp.
  157--165, 1952.

\bibitem{coulomb1773essai}
C.~A. Coulomb, ``Essai sur une application des regles de maximis et minimis a
  quelques problemes de statique relatifs a 1'architecture,'' \emph{Mem. Div.
  Sav. Acad.}, 1773.

\bibitem{sac}
T.~Haarnoja, A.~Zhou, P.~Abbeel, and S.~Levine, ``Soft actor-critic: Off-policy
  maximum entropy deep reinforcement learning with a stochastic actor,'' in
  \emph{International conference on machine learning}.\hskip 1em plus 0.5em
  minus 0.4em\relax PMLR, 2018, pp. 1861--1870.

\end{thebibliography}

\end{document}